\documentclass[runningheads,orivec]{llncs}
\usepackage[T1]{fontenc}
\usepackage{graphicx}
\usepackage{booktabs}
\usepackage{float}
\usepackage[misc]{ifsym}

\usepackage{amsmath,amsfonts}
\usepackage{array}
\usepackage[table]{xcolor}
\usepackage{tikz}
\newcommand{\R}{\mathbb{R}}
\newcommand{\calA}{\mathcal{A}}
\newcommand{\calL}{\mathcal{L}}
\newcommand{\calS}{\mathcal{S}}
\newcommand{\calD}{\mathcal{D}}
\newcommand{\zbase}{z^{0}}

\title{Local Evidence and Geometric Readout Repair in Trained GNNs}
\titlerunning{Local Evidence and Geometric Readout Repair}
\author{Nadi Tomeh \and Hugo Attali}
\authorrunning{N. Tomeh and H. Attali}
\institute{Universit\'e Sorbonne Paris Nord, CNRS,\\
Laboratoire d'Informatique de Paris Nord, LIPN,\\
F-93430 Villetaneuse, France\\
\email{\{tomeh,attali\}@lipn.fr}}

\begin{document}
\raggedbottom
\maketitle

\begin{abstract}
Many node-classification GNNs apply a linear classifier to a nonnegative
mixture of local messages.  An error can reflect either poor mixture weights or
a reachable logit set poorly positioned for the classifier.  We
separate these causes with an exact-mass linear program and two learned post-hoc
repairs.  Every reweighted prediction has an equivalent centered logit
translation, but only translations in a message-induced displacement set are
realizable by reweighting.  Across eight datasets, eight GNN backbones, and ten
splits, mean accuracy rises from $62.6\%$ for the frozen models to $63.8\%$ with
reweighting and $65.3\%$ with set-conditioned translation.  A parameter-matched
node-only translator reaches $64.6\%$, showing that translation explains most
of the gain while the message set supplies a smaller additional benefit.
Although oracle reweighting can correct many errors, label-free reweighting
captures little of this potential: local evidence is often present but hard to
select, and relaxing the evidence constraint is more effective than learning
within it.
\end{abstract}

\section{Introduction}

We study trained message-passing graph neural networks (GNNs) for transductive
node classification
\cite{scarselli2009gnn,gilmer2017mpnn,kipf2017gcn,velickovic2018gat,hamilton2017graphsage}.
Many end with a nonnegative weighted sum of final messages followed by a
linear classifier.  A local error can therefore arise because the coefficients
select an unhelpful mixture or because no admissible mixture of the final
messages supports the true class.  End-to-end accuracy conflates these causes.

If exact-mass reweighting makes the true class win, the fixed messages already
contain sufficient local evidence and only its selection needs repair.  If no
admissible reweighting succeeds, correction requires changing the messages or
another part of the model.  This distinction reveals whether a lightweight
readout repair can reuse existing evidence or must move beyond the
fixed-message polytope, a useful diagnosis when helpful and conflicting messages
coexist in heterophilous or noisy graphs
\cite{zhu2020h2gcn,chien2021gprgnn,lim2021linkx,platonov2023critical}.

We study the logit set induced by nonnegative reweighting with fixed branch
masses.  A linear program maximizes a proposed class's worst-rival margin; a
positive optimum certifies fixed-message support, and linear-program geometry
gives a compact solution.  With the GNN frozen, we train two label-free,
set-conditioned adapters from the same local information: one reweights within
the reachable set, while the other predicts a free centered logit translation.
The oracle measures evidence availability; comparing the nested repairs tests
label-free selection and the effect of relaxing the evidence constraint.  A
refitted head and a parameter-matched node-only translator separate these
effects from generic readout capacity and from translation that ignores the
message set.

\section{Fixed-Message Reachability}
\label{sec:geometry}

\subsection{Reachable logits and class margins}

Fix a node $v$ classified into one of $C$ classes.  Partition the adjustable
terms in its final aggregation into $B_v$ groups and write its frozen logits as
\begin{equation}
  \zbase_v=\beta_v+
  \sum_{g=1}^{B_v}\sum_{a\in\calA_{v,g}}
  \alpha_{v,a}\ell_{v,a}.
  \label{eq:readout}
\end{equation}
Here $\calA_{v,g}$ is group $g$, $\calA_v=\bigcup_g\calA_{v,g}$, and
$K_v=|\calA_v|$.  Message $h_{v,a}\in\R^{d_g}$ defines the neighbor-logit contribution
$\ell_{v,a}=W_gh_{v,a}\in\R^C$, where $W_g\in\R^{C\times d_g}$ is the frozen
branch transform and classifier.  The fixed term $\beta_v\in\R^C$ collects the classifier bias and
any non-adjustable root, residual, or branch-bias contribution, while
$\alpha_v\in\R_{\ge0}^{K_v}$ is the frozen coefficient vector and
\begin{equation}
  m_{v,g}=\sum_{a\in\calA_{v,g}}\alpha_{v,a}>0.
  \label{eq:group-mass}
\end{equation}
In GCN and SGC, $a$ indexes an incoming source and $\alpha_{v,a}$ is its
normalized propagation coefficient \cite{kipf2017gcn,wu2019sgc}.  In GAT and
GATv2, $a$ also identifies the head and $\alpha_{v,a}$ is the final attention
coefficient \cite{velickovic2018gat,brody2022attentive}.  GraphSAGE and
GraphConv expose their neighbor branch while keeping the root transform in
$\beta_v$ \cite{hamilton2017graphsage,morris2019weisfeiler}; MixHop and H2GCN
use one group per propagation stream
\cite{abuelhaija2019mixhop,zhu2020h2gcn}.  Appendix~\ref{app:setup} gives every
backbone's exact exposure.

Admissible coefficients preserve each group's frozen mass:
\begin{equation}
  \begin{aligned}
  \Delta_v
  &=\left\{\lambda\in\R_{\ge0}^{K_v}:
  \sum_{a\in\calA_{v,g}}\lambda_a=m_{v,g},\
  g=1,\ldots,B_v\right\},\\
  z_v(\lambda)
  &=\beta_v+\sum_{g=1}^{B_v}
  \sum_{a\in\calA_{v,g}}\lambda_a\ell_{v,a}.
  \end{aligned}
  \label{eq:mass-simplex}
\end{equation}
Every feasible $\lambda$ selects one logit vector; their set is the evidence
polytope $\calL_v=\{z_v(\lambda):\lambda\in\Delta_v\}$.  The frozen vector
$\alpha_v$ is feasible and selects the original output
$\zbase_v=z_v(\alpha_v)$.  Exact mass therefore isolates redistribution: it
cannot change a branch's scale or transfer mass across attention heads or
propagation streams.  For one group,
$\calL_v=\beta_v+m_{v,1}\operatorname{conv}\{\ell_{v,a}:a\in\calA_v\}$, so it
is the scaled convex hull of the individual neighbor-logit contributions.
Multiple groups add one point from each such hull.  Reweighting moves the
selected output within $\calL_v$ but cannot change the polytope itself.

The margin of class $c$ against rival $r$ is
\begin{equation}
  M_v^\lambda(c,r)=z_{v,c}(\lambda)-z_{v,r}(\lambda)
  =\beta_{v,c}-\beta_{v,r}+
  \sum_{g=1}^{B_v}\sum_{a\in\calA_{v,g}}
  \lambda_a(\ell_{v,a,c}-\ell_{v,a,r}),
  \quad r\ne c,
  \label{eq:pair-margin}
\end{equation}
Class $c$ wins exactly when $\min_{r\ne c}M_v^\lambda(c,r)\ge0$; its region is
$K_c=\{z\in\R^C:z_c\ge z_r\ \forall r\ne c\}$.  The minimum selects the
strongest rival at the chosen point, so a positive value means that $c$ beats
all alternatives, not only the frozen model's current prediction.

\subsection{Exact-mass oracle and compact solutions}
\label{sec:witnesses}

To test whether the fixed messages can support class $c$, we maximize its
worst-rival margin with the linear program (LP)
\begin{equation}
\begin{aligned}
  \gamma_v(c)=\max_{\lambda,\gamma}\quad &\gamma\\
  \mathrm{s.t.}\quad
  &\gamma\le M_v^\lambda(c,r) &&\forall r\ne c,\\
  &\lambda\in\Delta_v.&&
\end{aligned}
\label{eq:witness-lp}
\end{equation}
For fixed $\lambda$, the largest feasible $\gamma$ is its worst-rival margin,
so the outer maximization finds the best exact-mass weighting.  Positive
$\gamma$ gives a strict win, zero reaches only a decision boundary, and negative
$\gamma$ means every weighting loses to some rival.  For $c=y_v$, positivity
certifies coefficient-only correction; the label is used only in this oracle
analysis.

\begin{proposition}[Compact exact-mass optimum]
\label{prop:witness-region}
The evidence polytope $\calL_v$ intersects the interior of $K_c$ if and
only if $\gamma_v(c)>0$.  An optimal basic solution can be chosen with at most
$C+B_v-2$ active terms in total and at most $C-1$ active terms in any group.
\end{proposition}

The proof is in Appendix~\ref{app:proof}.  On a misclassified node,
$\gamma_v(y_v)>0$ identifies a selection failure; otherwise the true class
requires changing the messages or model.  The support bound is independent of
neighborhood size: a single-group readout needs at most $C-1$ terms, while each
additional positive-mass group must retain one.  A basic optimum thus gives a
small correcting set to inspect, but not necessarily a unique explanation or a
claim that the original GNN should aggregate only those terms.
Compactness is expected from LP geometry
\cite{boyd2004convex,bertsimas1997linear}; the substantive observation is
positive true-class margin, not sparsity alone.  Because the tested class is
supplied and the optimizer may be nonunique, the LP is an oracle diagnostic
rather than a prediction rule.  Support counts head- or stream-specific terms
separately.

\section{Learning Constrained and Free Readout Corrections}
\label{sec:learning}

We freeze the GNN and learn two adapters from the same local information.
Reweighting predicts coefficients and must select a point in $\calL_v$;
translation predicts a logit displacement that may leave this set.  Neither
changes the graph, messages, or classifier, so their comparison isolates the
cost of restricting a correction to combinations of the frozen messages.

\begin{figure}[H]
\centering
\begin{tikzpicture}[x=0.78cm,y=0.78cm]
  \path[use as bounding box] (-5.0,-2.0) rectangle (4.8,2.8);
  \fill[red!9]
    (-5.0,-2.0)--(4.8,-2.0)--(4.8,0.59)--(-5.0,-1.37)--cycle;
  \fill[blue!9]
    (-5.0,-1.37)--(-5.0,2.8)--(4.8,2.8)--(4.8,0.59)--cycle;
  \draw[black!38,line width=0.45pt] (-5.0,-1.37)--(4.8,0.59);
  \node[blue!55!black] at (-3.0,2.15) {class $c$ region};
  \node[red!60!black] at (3.45,-0.85) {rival region};

  \filldraw[fill=blue!28,fill opacity=0.24,draw=blue!65,line width=0.8pt]
    (-3.00,-1.10)--(-2.50,0.55)--(-1.10,0.80)--(-0.20,0.10)--(-0.70,-1.45)--cycle;
  \foreach \p in {(-3.00,-1.10),(-2.50,0.55),(-1.10,0.80),(-0.20,0.10),(-0.70,-1.45)}
    \fill[blue!65!black] \p circle (1.15pt);
  \node[blue!65!black,anchor=north] at (-2.05,-1.50) {$\calL_v$};

  \fill[black] (-0.90,-0.95) circle (1.6pt);
  \node[black!75,anchor=north west] at (-0.76,-1.02) {$z_v(\alpha_v)$};
  \fill[orange!85!black] (-1.70,0.05) circle (1.7pt);
  \draw[->,orange!85!black,line width=1.0pt] (-0.95,-0.88)--(-1.62,-0.02);
  \node[orange!70!black,anchor=east] at (-3.15,0.20)
    {reweight $=q_v^\lambda$};

  \draw[->,teal!70!black,line width=1.1pt] (-0.90,-0.95)--(1.90,0.45)
    node[midway,below] {free $q_v$};
  \draw[teal!70!black,dashed,line width=0.9pt]
    (-0.20,0.30)--(0.30,1.95)--(1.70,2.20)--(2.60,1.50)--(2.10,-0.05)--cycle;
  \fill[teal!70!black] (1.90,0.45) circle (1.6pt);
  \node[teal!55!black,anchor=south] at (1.25,2.28) {$\calL_v+q_v$};
  \node[teal!60!black,anchor=west] at (2.78,1.70) {translate};
\end{tikzpicture}
\caption{Reweighting remains in $\calL_v$ and has an equivalent centered
translation $q_v^\lambda$.  A free $q_v$ may lie outside the feasible
displacements and shifts the whole set (dashed) without changing its shape.}
\label{fig:repair-geometry}
\end{figure}
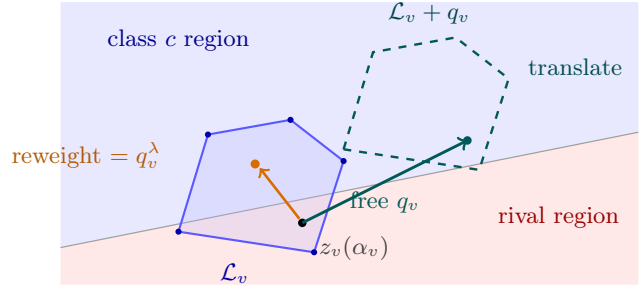

\subsection{Shared local encoding}

Both adapters receive the same final-stage set $\calS_v$: the adjustable
messages in $\calA_v$ plus exposed fixed terms such as a separate root
contribution.  Thus $\calA_v\subseteq\calS_v$.  The full set describes the
alternatives around $v$; fixed terms provide context but are not reweighted.

For each $a\in\calS_v$, we use the feature vector
\begin{equation}
  x_{v,a}=[h_v,h_{v,a},h_{v,a}-h_v,\zbase_v,\zbase_a,
  \zbase_a-\zbase_v,\cos(h_v,h_{v,a}),
  \|h_{v,a}-h_v\|_2]\in\R^{3d+3C+2},
  \label{eq:pair-feature}
\end{equation}
where $\zbase_a\in\R^C$ is the frozen source-node logit vector.  These features
compare recipient and message in both spaces: relative representation features
describe compatibility with $v$, while frozen logits expose class evidence.

Each term retains its fixed readout weight $\omega_{v,a}\geq0$:
$\omega_{v,a}=\alpha_{v,a}$ for adjustable terms, while an exposed root has
weight one.  These inputs weight the context summary, not the predicted
coefficients.  With $\bar m_v=\sum_{a\in\calS_v}\omega_{v,a}$ and branch index
$k\in\{\mathrm R,\mathrm T\}$, where R denotes reweighting and T translation,
each adapter uses
\begin{equation}
  p_v^k=\frac{1}{\bar m_v}
  \sum_{a\in\calS_v}\omega_{v,a}\psi_k(x_{v,a}),
  \qquad
  c_v^k=[h_v,\zbase_v,p_v^k,\bar m_v].
  \label{eq:shared-context}
\end{equation}
where $p_v^k\in\R^{d_p}$ and $c_v^k\in\R^{d+C+d_p+1}$.  The separately
parameterized encoders are permutation invariant and handle variable set sizes;
$h_v$, $\zbase_v$, and $\bar m_v$ retain recipient state and scale.  Thus
$p_v^k$ summarizes the weighted set, and $c_v^k$ augments it with the recipient.

\subsection{Readout repair variants}

\paragraph{Evidence-constrained correction.}
The reweighting branch scores each adjustable message in context and tilts its
frozen coefficient; groupwise normalization preserves the original mass:
\begin{equation}
  \begin{aligned}
  s_{v,a}&=s_\theta([x_{v,a},c_v^{\mathrm R}])\in\R,\\
  \lambda^\theta_{v,a}
  &=m_{v,g}\frac{\alpha_{v,a}\exp s_{v,a}}
  {\sum_{b\in\calA_{v,g}}\alpha_{v,b}\exp s_{v,b}},
  \qquad a\in\calA_{v,g}.
  \end{aligned}
  \label{eq:learned-coefficients}
\end{equation}
The exponential makes every coefficient nonnegative, and the denominator
enforces $\sum_{a\in\calA_{v,g}}\lambda_{v,a}^\theta=m_{v,g}$ exactly.  Thus
$\lambda_v^\theta\in\Delta_v$: the adapter changes the selected point but not
the messages, support, or polytope.  Zero initialization gives
$\lambda_v^\theta=\alpha_v$, and
$z_v^{\mathrm R}=z_v(\lambda_v^\theta)$.  Because the update is multiplicative,
a term with zero frozen coefficient remains unavailable; the adapter neither
adds edges nor enlarges the original support.

\paragraph{Free logit correction.}
The translation branch maps $c_v^{\mathrm T}$ to a node-specific shift.  We
remove the common-logit component, which does not affect classification:
\begin{equation}
  \bar q_v=\rho_\phi(c_v^{\mathrm T})\in\R^C,
  \qquad
  q_v=\bar q_v-\frac{{\bf1}^\top\bar q_v}{C}{\bf1},
  \qquad z_v^{\mathrm T}=\zbase_v+q_v.
  \label{eq:translation-adapter}
\end{equation}
Centering removes the unidentifiable all-ones direction of softmax logits.
Adding the same $q_v$ to every $z_v(\lambda)$ then translates $\calL_v$ rigidly,
without changing its shape or the relative position selected by $\alpha_v$.
``Free'' means that $q_v$ need not be realizable by reweighting; its magnitude
remains bounded in the implementation.  This is a residual correction at the
frozen readout, not a modification of individual messages or graph edges.

\paragraph{Why reweighting is contained in translation.}
Let $P=I-\frac{1}{C}{\bf1}{\bf1}^\top\in\R^{C\times C}$ center a logit vector and define the
centered reweighting-displacement set
\begin{equation}
  \calD_v=P(\calL_v-\zbase_v)
  =\{P(z_v(\lambda)-\zbase_v):\lambda\in\Delta_v\}.
  \label{eq:displacement-set}
\end{equation}

\begin{proposition}[Reweighting is contained in translation]
\label{prop:pointwise-containment}
For every $\lambda\in\Delta_v$, the centered translation
$q_v^\lambda=P(z_v(\lambda)-\zbase_v)$ makes $\zbase_v+q_v^\lambda$ produce the
same softmax probabilities and class prediction as $z_v(\lambda)$.  Conversely, a centered translation
$q_v$ is equivalent to an exact-mass reweighting if and only if
$q_v\in\calD_v$.
\end{proposition}

For a chosen $\lambda$, adding $q_v^\lambda$ to $\zbase_v$ recovers
$z_v(\lambda)$ up to a common offset, hence gives the same softmax.  Conversely,
$\calD_v$ is exactly the set of such feasible displacements.  The proposition
is pointwise, not an equivalence of learned adapters: reweighting must infer an
output in $\calD_v$, whereas translation may predict any bounded centered
$q_v$.  Since their inputs are the same, an accuracy gap isolates this output
constraint rather than a difference in observed information.

\subsection{Training the adapters}

Each adapter minimizes mean training-node cross-entropy, updating only $\theta$
or $\phi$ while the graph, backbone, messages, support, and classifier remain
fixed.  ``Label-free'' refers to prediction: labels supervise training, but the
target class is not an input, and test-time repair uses only $\calS_v$.  Both
start as exact no-ops, so validation selection may retain the frozen rule when a
repair does not help.  Appendix~\ref{app:setup} gives optimization details.

\subsection{Relation to existing methods}

Graph attention and adaptive propagation learn coefficients jointly with the
representations \cite{velickovic2018gat,brody2022attentive,chien2021gprgnn},
while sparse attention constrains their geometry
\cite{martins2016sparsemax,niculae2017regularized}.  We instead freeze the GNN
and train small post-hoc modules, following adapter and side-network paradigms
\cite{houlsby2019adapters,zhang2020sidetuning}.  Our adapters use a weighted Deep Sets
encoder \cite{zaheer2017deepsets} at the final readout.  Unlike calibration
\cite{guo2017calibration,hsu2022miscalibrated}, the objective is error
correction; unlike edge-conditioned or feature-wise message modulation
\cite{simonovsky2017ecc,brockschmidt2020gnnfilm}, earlier GNN layers remain
unchanged.

Compact oracle supports are related to GNN and counterfactual explanations
\cite{ying2019gnnexplainer,luo2020pgexplainer,lucic2022cfgnnexplainer}, but our
LP tests class reachability under fractional exact-mass reweighting rather than
explaining a fixed prediction or editing the graph.  Graph diffusion and
Correct-and-Smooth act through graph-wide prediction propagation
\cite{klicpera2019appnp,gasteiger2019diffusion,huang2021correctsmooth}; our
repairs remain local to the trained final message set.

\section{Experiments}
\label{sec:experiments}

We evaluate eight datasets and eight backbones over ten transductive node-
classification splits.  Validation accuracy selects each backbone checkpoint;
the graph, messages, coefficients, and classifier are then frozen for all
diagnostics and repairs.  Table cells are ten-split means, and macro averages
weight the $64$ dataset--backbone cases equally.  Appendix~\ref{app:setup} gives
the datasets, architectures, optimization, and batching details.

\subsection{Oracle reachability of frozen errors}

For every validation and test node, we solve Eq.~\eqref{eq:witness-lp} for the
true class with Gurobi \cite{gurobi2026}, preserving every group mass.  Let
$\widehat y_v^0=\arg\max_c \zbase_{v,c}$.  Table~\ref{tab:frozen-lp} reports on
test nodes
\begin{equation}
\begin{aligned}
 \mathrm{TCReach}
   &=\Pr[\gamma_v(y_v)>0],\\
 \mathrm{ErrReach}
   &=\Pr[\gamma_v(y_v)>0\mid \widehat y_v^0\ne y_v],\\
 \mathrm{Support}
   &=\mathbb E[\|\lambda_v^\star\|_0\mid \gamma_v(y_v)>0].
\end{aligned}
\label{eq:oracle-metrics}
\end{equation}
TC reach covers all nodes, Err. reach conditions on frozen-model mistakes, and
Support counts the active coefficients in one positive-margin basic optimum.
The test label selects the LP target, so these are oracle diagnostics rather
than predictions.  A nonpositive score rules out correction within this frozen
final readout; it does not rule out useful information elsewhere in the graph.
Appendix~\ref{app:setup} gives numerical thresholds and solver details.

\begin{table}[H]
\centering
\caption{Exact-mass frozen-message oracle diagnostics over $n=10$ splits, with backbones in rows and datasets in columns. TC reach and Err. reach are percentages; Support is the mean number of active coefficients in positive true-class solutions.}
\label{tab:frozen-lp}
\scriptsize
\setlength{\tabcolsep}{2.3pt}
\renewcommand{\arraystretch}{0.9}
\begin{tabular}{@{}l*{9}{r}@{}}
\toprule
\rowcolor{black!7}\multicolumn{10}{@{}l}{\textbf{(a) True-class reach (\%)}} \\
Model & Actor & Amazon & Cham. & Cite. & Cora & PubMed & Roman & Squir. & Avg. \\
\midrule
GCN & 62.2 & 69.4 & 94.0 & 80.8 & 95.3 & 95.6 & 78.8 & 94.6 & 83.8 \\
GAT & 59.3 & 62.2 & 92.1 & 81.1 & 94.5 & 94.1 & 70.1 & 91.0 & 80.5 \\
SAGE & 52.2 & 68.4 & 78.4 & 79.7 & 94.1 & 93.2 & 92.8 & 66.2 & 78.1 \\
GATv2 & 62.7 & 66.8 & 92.9 & 80.7 & 94.6 & 94.1 & 83.3 & 92.2 & 83.4 \\
GraphConv & 56.4 & 65.5 & 72.4 & 80.7 & 93.9 & 94.0 & 93.8 & 46.5 & 75.4 \\
SGC & 94.0 & 87.8 & 99.6 & 89.2 & 99.1 & 99.6 & 96.6 & 100.0 & 95.7 \\
MixHop & 72.2 & 87.9 & 91.0 & 86.5 & 97.9 & 97.6 & 97.7 & 68.8 & 87.4 \\
H2GCN & 79.7 & 88.5 & 96.2 & 83.3 & 96.4 & 96.8 & 96.1 & 91.0 & 91.0 \\
\cmidrule(lr){1-10}
Avg. & 67.3 & 74.6 & 89.6 & 82.7 & 95.7 & 95.6 & 88.7 & 81.3 & 84.4 \\
\midrule
\rowcolor{black!7}\multicolumn{10}{@{}l}{\textbf{(b) Error reach (\%)}} \\
Model & Actor & Amazon & Cham. & Cite. & Cora & PubMed & Roman & Squir. & Avg. \\
\midrule
GCN & 45.8 & 43.8 & 80.8 & 20.4 & 60.4 & 59.2 & 54.1 & 88.6 & 56.6 \\
GAT & 42.2 & 31.2 & 77.5 & 22.8 & 55.4 & 53.0 & 43.2 & 82.0 & 50.9 \\
SAGE & 26.1 & 39.7 & 45.0 & 17.8 & 51.7 & 32.3 & 65.6 & 41.2 & 39.9 \\
GATv2 & 46.8 & 37.2 & 78.8 & 22.0 & 56.3 & 51.3 & 53.7 & 83.9 & 53.7 \\
GraphConv & 37.7 & 38.9 & 33.6 & 23.2 & 53.9 & 43.8 & 69.7 & 25.7 & 40.8 \\
SGC & 91.5 & 78.8 & 98.9 & 54.7 & 92.8 & 96.3 & 93.6 & 99.9 & 88.3 \\
MixHop & 57.5 & 79.1 & 79.7 & 42.3 & 84.4 & 77.7 & 90.0 & 51.6 & 70.3 \\
H2GCN & 68.7 & 80.0 & 90.5 & 32.6 & 72.9 & 69.5 & 83.6 & 84.7 & 72.8 \\
\cmidrule(lr){1-10}
Avg. & 52.0 & 53.6 & 73.1 & 29.5 & 66.0 & 60.4 & 69.2 & 69.7 & 59.2 \\
\midrule
\rowcolor{black!7}\multicolumn{10}{@{}l}{\textbf{(c) Active support}} \\
Model & Actor & Amazon & Cham. & Cite. & Cora & PubMed & Roman & Squir. & Avg. \\
\midrule
GCN & 1.40 & 1.22 & 1.65 & 1.07 & 1.12 & 1.11 & 1.95 & 1.79 & 1.41 \\
GAT & 1.40 & 1.22 & 1.56 & 1.11 & 1.19 & 1.10 & 1.82 & 1.80 & 1.40 \\
SAGE & 1.25 & 1.30 & 1.33 & 1.08 & 1.12 & 1.05 & 1.61 & 1.33 & 1.26 \\
GATv2 & 1.43 & 1.25 & 1.52 & 1.11 & 1.18 & 1.09 & 1.69 & 1.71 & 1.37 \\
GraphConv & 1.50 & 1.27 & 1.38 & 1.09 & 1.13 & 1.13 & 1.81 & 1.29 & 1.32 \\
SGC & 2.20 & 1.60 & 2.24 & 1.48 & 1.79 & 1.26 & 2.31 & 2.83 & 1.96 \\
MixHop & 3.52 & 3.64 & 3.65 & 3.27 & 3.52 & 3.10 & 4.41 & 3.60 & 3.59 \\
H2GCN & 3.68 & 3.71 & 3.90 & 3.21 & 3.43 & 3.11 & 4.15 & 3.86 & 3.63 \\
\cmidrule(lr){1-10}
Avg. & 2.05 & 1.90 & 2.15 & 1.68 & 1.81 & 1.62 & 2.47 & 2.27 & 1.99 \\
\bottomrule
\end{tabular}
\end{table}

Across the $64$ dataset--backbone cases, macro true-class reach is $84.4\%$ and
error reach is $59.2\%$.  Thus many errors select the wrong point from an
adequate local set, while about two fifths are not strictly correctable under
the exact-mass constraints.  Error reach ranges from $39.9\%$ for GraphSAGE to
$88.3\%$ for SGC and from $29.5\%$ on Citeseer to $73.1\%$ on Chameleon.  A
positive solution uses $1.99$ active terms on average; MixHop and H2GCN use more
because every stream retains a term.  This compact support aids inspection, but
reachability is a label-aware evidence diagnostic, not a deployable predictor.

\subsection{Accuracy and controls for the learned repairs}

Table~\ref{tab:repair-results} compares the frozen model with learned
reweighting and set-conditioned translation for every dataset and backbone.

\begin{table}[H]
\centering
\caption{Mean test accuracy (\%) over ten splits. Rows are backbones and columns are datasets; bold marks the best learned repair when it improves on Frozen.}
\label{tab:repair-results}
\scriptsize
\setlength{\tabcolsep}{2.3pt}
\renewcommand{\arraystretch}{0.80}
\begin{tabular}{@{}l*{9}{r}@{}}
\toprule
\rowcolor{black!7}\multicolumn{10}{@{}l}{\textbf{(a) Frozen}} \\
Model & Actor & Amazon & Cham. & Cite. & Cora & PubMed & Roman & Squir. & Avg. \\
\midrule
GCN & 30.2 & 45.6 & 68.4 & 75.9 & 88.1 & 89.2 & 53.8 & 52.7 & 63.0 \\
GAT & 29.5 & 45.0 & 64.9 & 75.4 & 87.8 & 87.5 & 47.4 & 50.3 & 61.0 \\
SAGE & 35.3 & 47.6 & 60.8 & 75.3 & 87.8 & 89.9 & 79.2 & 42.4 & 64.8 \\
GATv2 & 29.8 & 47.1 & 66.6 & 75.2 & 87.7 & 87.8 & 64.0 & 51.7 & 63.7 \\
GraphConv & 30.2 & 43.7 & 58.4 & 74.9 & 86.9 & 89.3 & 79.6 & 28.4 & 61.4 \\
SGC & 29.6 & 42.4 & 67.5 & 76.2 & 88.6 & 88.1 & 46.5 & 50.0 & 61.1 \\
MixHop & 34.5 & 42.0 & 55.4 & 76.7 & 86.6 & 89.4 & 76.9 & 35.7 & 62.1 \\
H2GCN & 35.2 & 42.5 & 60.3 & 75.2 & 86.8 & 89.4 & 76.3 & 41.0 & 63.3 \\
\cmidrule(lr){1-10}
Avg. & 31.8 & 44.5 & 62.8 & 75.6 & 87.5 & 88.8 & 65.5 & 44.0 & 62.6 \\
\midrule
\rowcolor{black!7}\multicolumn{10}{@{}l}{\textbf{(b) Reweight}} \\
Model & Actor & Amazon & Cham. & Cite. & Cora & PubMed & Roman & Squir. & Avg. \\
\midrule
GCN & 30.5 & 45.6 & \textbf{68.5} & \textbf{76.0} & \textbf{88.1} & 89.3 & 59.6 & 54.2 & 64.0 \\
GAT & \textbf{30.0} & 45.1 & 64.9 & 75.6 & 87.8 & 88.5 & 49.9 & 52.3 & 61.8 \\
SAGE & \textbf{35.4} & 47.6 & 60.7 & \textbf{75.5} & 87.9 & 90.0 & \textbf{79.8} & \textbf{42.6} & 64.9 \\
GATv2 & 30.0 & 47.0 & 66.8 & \textbf{75.3} & \textbf{87.8} & 88.4 & 69.6 & 52.1 & 64.6 \\
GraphConv & 30.2 & 43.7 & 58.4 & 74.8 & 86.9 & 89.3 & \textbf{80.5} & 29.1 & 61.6 \\
SGC & 33.1 & 48.9 & \textbf{67.8} & 76.1 & \textbf{88.6} & 89.6 & 67.5 & \textbf{51.1} & 65.3 \\
MixHop & 34.4 & 46.5 & \textbf{56.8} & 76.2 & \textbf{87.4} & \textbf{89.8} & 79.5 & \textbf{37.2} & 63.5 \\
H2GCN & 35.0 & 47.3 & \textbf{61.9} & 75.2 & 86.8 & \textbf{89.5} & 78.1 & \textbf{42.6} & 64.5 \\
\cmidrule(lr){1-10}
Avg. & 32.3 & 46.5 & 63.2 & 75.6 & 87.7 & 89.3 & 70.6 & 45.2 & 63.8 \\
\midrule
\rowcolor{black!7}\multicolumn{10}{@{}l}{\textbf{(c) Set translation}} \\
Model & Actor & Amazon & Cham. & Cite. & Cora & PubMed & Roman & Squir. & Avg. \\
\midrule
GCN & \textbf{30.8} & \textbf{45.8} & 68.4 & 75.9 & 88.1 & \textbf{89.4} & \textbf{76.3} & \textbf{57.3} & 66.5 \\
GAT & 30.0 & \textbf{46.4} & \textbf{67.3} & \textbf{75.6} & \textbf{88.0} & \textbf{88.8} & \textbf{76.2} & \textbf{54.7} & 65.9 \\
SAGE & 35.4 & \textbf{47.7} & \textbf{61.1} & 75.3 & \textbf{88.0} & \textbf{90.0} & 79.1 & 42.4 & 64.9 \\
GATv2 & \textbf{30.2} & \textbf{47.1} & \textbf{67.1} & 75.0 & 87.4 & \textbf{88.7} & \textbf{80.7} & \textbf{53.3} & 66.2 \\
GraphConv & \textbf{30.5} & \textbf{43.8} & \textbf{62.1} & 74.8 & \textbf{87.1} & \textbf{89.3} & 79.8 & \textbf{38.0} & 63.2 \\
SGC & \textbf{35.3} & \textbf{49.4} & 67.8 & 76.1 & 88.3 & \textbf{89.7} & \textbf{78.7} & 50.3 & 67.0 \\
MixHop & \textbf{34.8} & \textbf{49.2} & 56.0 & \textbf{76.8} & 86.5 & 89.8 & \textbf{81.0} & 36.0 & 63.7 \\
H2GCN & \textbf{35.5} & \textbf{50.0} & 60.9 & \textbf{75.3} & \textbf{86.9} & 89.3 & \textbf{79.0} & 41.2 & 64.8 \\
\cmidrule(lr){1-10}
Avg. & 32.8 & 47.4 & 63.8 & 75.6 & 87.5 & 89.4 & 78.9 & 46.6 & \textbf{65.3} \\
\bottomrule
\end{tabular}
\end{table}

To test whether these gains only reflect additional post-hoc capacity,
Table~\ref{tab:translation-controls} adds two controls.  Refit head learns a new
affine classifier on the frozen final embeddings and tests whether moving the
linear decision boundary is sufficient.  Node translation keeps the original
classifier and predicts the same bounded centered-logit correction as set
translation, but from $[h_v,\zbase_v]$ alone.  Its hidden width is chosen
separately in every case to match the set translator's parameter count within
$1\%$.  The comparison between the two translators therefore tests the value of
the message-set input at matched output space and capacity.  Appendix~
\ref{app:controls} gives the architectures, optimization protocol, and complete
$64$-case results.

\begin{table}[H]
\centering
\caption{Capacity and information controls. Test accuracy is the macro mean over all $64$ dataset--backbone cases; every case averages ten splits. The node translator is parameter matched to the set translator within $1\%$ in every case.}
\label{tab:translation-controls}
\small
\setlength{\tabcolsep}{5.2pt}
\begin{tabular}{@{}lccrr@{}}
\toprule
Method & Learned output & Message set & Accuracy & $\Delta$ Frozen \\
\midrule
Frozen & none & no & 62.6 & -- \\
Refit head & linear head & no & 62.1 & -0.5 \\
Node translation & centered shift & no & 64.6 & +2.1 \\
Set reweight & coefficients & yes & 63.8 & +1.2 \\
Set translation & centered shift & yes & \textbf{65.3} & \textbf{+2.7} \\
\bottomrule
\end{tabular}
\end{table}

The $64$-case mean rises from $62.6\%$ to $63.8\%$ with reweighting, far below
the oracle opportunity in Table~\ref{tab:frozen-lp}: the oracle sees the target
class, whereas the adapter must infer weights without it.

Set translation reaches $65.3\%$ and improves on Frozen in $53$ cases.  Gains
are largest on Roman-empire ($+13.4$ points), Amazon-ratings ($+2.9$), and
Squirrel ($+2.6$), but negligible on the citation graphs.  Its $1.5$-point
advantage over reweighting reflects freedom to leave $\calD_v$; inside
$\calD_v$, the two corrections are pointwise equivalent
(Proposition~\ref{prop:pointwise-containment}).

The controls refine this conclusion.  Refitting the head decreases the macro
mean by $0.5$ points, while node translation gains $2.1$ points despite having
no message-set input.  Translation itself therefore explains most of the
improvement.  Set conditioning adds another $0.63$ points and wins in $40$ of
$64$ cases under parameter matching.  This increment is concentrated but not
exclusive to Roman-empire: it is $+3.38$ points there, $+0.96$ on Squirrel, and
$+0.24$ over the $56$ non-Roman cases.  Thus the full local set can help choose
the correction direction, but it is not required for every graph or backbone.

\section{Conclusion}

The exact-mass oracle shows that many errors have a compact true-class
combination, but learned reweighting captures little of this opportunity.  Free
translation performs better because it can express corrections outside the
fixed-message polytope.  The controls show that a node-only translation already
captures most of this benefit, while the full message set provides a smaller
additional gain at matched capacity.  Local evidence therefore has two roles:
it defines the reweighting diagnostic and can improve how a free correction
direction is selected.  Future work may learn stable near-optimal sets rather
than one LP solution and test larger structured relaxations.

\bibliographystyle{splncs04}
\bibliography{references}

\appendix

\section{Proof of Proposition~\ref{prop:witness-region}}
\label{app:proof}

For any $\lambda\in\Delta_v$, the $r$th class-$c$ margin coordinate is
\begin{equation}
  z_{v,c}(\lambda)-z_{v,r}(\lambda)=M_v^\lambda(c,r),
  \qquad r\ne c.
  \label{eq:proof-margin}
\end{equation}
Consequently, $\gamma_v(c)>0$ holds exactly when one feasible logit has every
class-$c$ margin positive, which is equivalent to intersection with the
interior of $K_c$.

The feasible coefficient set is compact, so the LP has an optimal basic
solution.  The full LP has $K_v+1$ variables: $K_v$ coefficients and
$\gamma$.  If $k$ coefficients are positive, the other $K_v-k$
nonnegativity constraints are active.  The $B_v$ exact-mass equalities are
linearly independent because the groups are disjoint and have positive mass,
and there are at most $C-1$ active rival-margin constraints.  A basic solution
requires $K_v+1$ linearly independent active constraints, hence
\begin{equation}
  K_v+1\le (K_v-k)+B_v+(C-1),\qquad\text{hence}\qquad
  k\le C+B_v-2.
  \label{eq:basic-support-bound}
\end{equation}
Every positive-mass group contains at least one active coefficient.  Removing
the mandatory active term from each of the other $B_v-1$ groups yields
$k_g\le k-(B_v-1)\le C-1$ for each group $g$.  This proves both support bounds.

\section{Experimental Details}
\label{app:setup}

\subsection{Datasets, preprocessing, and splits}

Table~\ref{tab:dataset-details} summarizes the eight node-classification
datasets.  Cora, Citeseer, and PubMed follow the Planetoid collection
\cite{yang2016planetoid}; Actor, Chameleon, and Squirrel follow the Geom-GCN
collection \cite{pei2020geomgcn}; and Amazon-ratings and Roman-empire come from
the Heterophilous Graph Benchmark \cite{platonov2023critical}.  $E$ counts
directed entries after preprocessing.  For every dataset,
we cast the provided node features to floating point without further
normalization, symmetrize the graph, coalesce duplicate directed entries, and
add one self-loop per node.  The resulting graph and all node features are
available during transductive training and evaluation.

\begin{table}[H]
\centering
\caption{Dataset statistics after graph preprocessing.  Split percentages are
train/validation/test; ``provided'' denotes the ten masks distributed with the
dataset.}
\label{tab:dataset-details}
\small
\setlength{\tabcolsep}{4.5pt}
\begin{tabular}{@{}lrrrrl@{}}
\toprule
Dataset & $N$ & $E$ & Features & Classes & Ten-split protocol \\
\midrule
Actor          & 7,600  & 61,011  & 932   & 5  & provided, 48/32/20 \\
Amazon-ratings & 24,492 & 210,592 & 300   & 5  & provided, 50/25/25 \\
Chameleon      & 2,277  & 65,069  & 2,325 & 5  & provided, 48/32/20 \\
Citeseer       & 3,327  & 12,431  & 3,703 & 6  & random, 60/20/20 \\
Cora           & 2,708  & 13,264  & 1,433 & 7  & random, 60/20/20 \\
PubMed         & 19,717 & 108,365 & 500   & 3  & random, 60/20/20 \\
Roman-empire   & 22,662 & 88,516  & 300   & 18 & provided, 50/25/25 \\
Squirrel       & 5,201  & 402,047 & 2,089 & 5  & provided, 48/32/20 \\
\bottomrule
\end{tabular}
\end{table}

For Actor, Amazon-ratings, Chameleon, Roman-empire, and Squirrel, we use
provided split columns $0$--$9$.  Citeseer, Cora, and PubMed provide only one
standard mask in this data interface, so we create ten class-stratified splits
with seeds $100$--$109$ and ratios $60/20/20$.
Every reported entry averages these ten splits, which addresses the known
split sensitivity of GNN evaluation
\cite{shchur2018pitfalls,platonov2023critical}.  Model and adapter initialization
uses seed $0$ independently on every split.  Training labels fit the backbone
and adapter, validation accuracy selects checkpoints, and test labels are used
only for final accuracy and the explicitly labeled oracle analysis.

\subsection{Backbone architectures and training}

We evaluate GCN \cite{kipf2017gcn}, GAT \cite{velickovic2018gat}, GraphSAGE
\cite{hamilton2017graphsage}, GATv2 \cite{brody2022attentive}, GraphConv
\cite{morris2019weisfeiler}, SGC \cite{wu2019sgc}, MixHop
\cite{abuelhaija2019mixhop}, and H2GCN \cite{zhu2020h2gcn}.  All use hidden
width $128$.  GCN, GAT, GraphSAGE, GATv2, and GraphConv have two
message-passing layers.  GAT and GATv2 use eight concatenated heads in the first
layer and one non-concatenated output head.  SGC is one linear filter with two
propagation steps, MixHop is one layer with powers $\{0,1,2\}$, and our
H2GCN-style layer concatenates root, one-hop mean, and two-hop mean channels.
ReLU and dropout are applied between message-passing layers; attention dropout
is also applied inside GAT and GATv2.

All backbones minimize training-node cross-entropy with Adam
\cite{kingma2015adam}, learning rate $10^{-2}$, weight decay
$5\times10^{-4}$, and gradient clipping at $5.0$.  The dropout argument is
$0.5$.  GCN, GAT, and GraphSAGE run for $200$ epochs.  The five added
backbones run for at most $1000$ epochs and stop after $100$ epochs without a
strict validation-accuracy improvement.  In both cases, the checkpoint with
the highest validation accuracy is retained.  Backbone parameters, including
the classifier, are subsequently frozen.

\subsection{Final-stage term exposure}

Table~\ref{tab:readout-exposure} specifies how Eq.~\eqref{eq:readout} is
constructed for each architecture.  A higher-order source--recipient entry is
the corresponding nonzero entry after sparse matrix multiplication; multiple
paths to the same source are therefore combined into one coefficient.  The
translation adapter additionally receives fixed root terms for GraphSAGE and
GraphConv as context with weight one, although those terms remain in
$\beta_v$ and are not reweighted.

\begin{table}[H]
\centering
\caption{Architecture-faithful exposure of the final aggregation.  Every group
preserves its own frozen coefficient mass.}
\label{tab:readout-exposure}
\footnotesize
\setlength{\tabcolsep}{3.2pt}
\begin{tabular}{@{}>{\raggedright\arraybackslash}p{1.90cm}
                    >{\raggedright\arraybackslash}p{3.95cm}
                    >{\centering\arraybackslash}p{1.35cm}
                    >{\raggedright\arraybackslash}p{3.85cm}@{}}
\toprule
Backbone & Adjustable term and frozen coefficient & Groups & Fixed contribution in $\beta_v$ \\
\midrule
GCN & Normalized final-layer edge/self-loop term; $\alpha_{v,a}=\widehat A_{va}$ & 1 & Convolution and classifier biases \\
GAT/GATv2 & Final attention edge--head copy; $\alpha_{v,a}$ is its attention coefficient & 1 & Residual, layer bias, and classifier bias \\
GraphSAGE & Neighbor/self-loop term; $\alpha_{v,a}=1/\deg(v)$ & 1 & Separate root transform, neighbor-branch bias, and classifier bias \\
GraphConv & Neighbor/self-loop term; $\alpha_{v,a}=1$ & 1 & Separate root transform, neighbor-branch bias, and classifier bias \\
SGC & Nonzero entry of $\widehat A^2$ & 1 & Linear and classifier biases \\
MixHop & Nonzero entry of $\widehat A^p$, $p\in\{0,1,2\}$ & 3 & Concatenated-stream and classifier biases \\
H2GCN & Nonzero entry of $I$, $P$, or $P^2$, with $P$ the no-self mean operator & 3 & Concatenated-stream and classifier biases \\
\bottomrule
\end{tabular}
\end{table}

\subsection{Repair adapters and optimization}

The term encoder $\psi_k$, reweighting scorer $s_\theta$, and translation head
$\rho_\phi$ are two-layer MLPs with hidden width $H=128$, ReLU, and dropout
$0.2$; the pooled representation has dimension $d_p=128$.  The final affine
layer of each output head is initialized to zero.  If $\widehat s_{v,a}$ and
$\widehat q_v$ denote its unbounded outputs, the implementation uses
$s_{v,a}=4\tanh(\widehat s_{v,a})$ in
Eq.~\eqref{eq:learned-coefficients} and centers
$4\tanh(\widehat q_v)$ as in Eq.~\eqref{eq:translation-adapter}.  Reweighting
normalizes separately within every group, so its output satisfies the exact
mass constraints throughout training.

The two adapters use the same width and feature definition but are not
parameter matched.  For pair-feature dimension $d_x=3d+3C+2$, define the
parameter count of a two-layer width-$H$ MLP by
$F(i,o)=iH+H+Ho+o$.  Translation has
$F(d_x,H)+F(d+C+H+1,C)$ parameters, whereas reweighting has
$F(d_x,H)+F(d_x+d+C+H+1,1)$.  Across the reported cases these range from
$101$K--$1.93$M and $150$K--$3.36$M parameters, respectively.  We do not tune
depth or width separately, so the comparison concerns these common
architectures rather than capacity-matched optima.

Both adapters minimize mean training-node cross-entropy with AdamW
\cite{loshchilov2019adamw}, weight decay $10^{-4}$, and gradient clipping at
$5.0$.  Translation uses learning rate $10^{-3}$, at most $500$ epochs, and
patience $80$; reweighting uses learning rate $5\times10^{-4}$, at most $300$
epochs, and patience $60$.  Validation accuracy selects the checkpoint, with
the frozen prediction retained as epoch zero.  Zero output makes translation
an exact no-op and returns the original coefficient rule for reweighting; its
separate sparse accumulation may differ from the saved frozen logits only by
floating-point accumulation order.

Edge terms are evaluated in chunks without truncating any local set: chunks
contain $65{,}536$ terms on V100 cases and $32{,}768$ terms on A100 cases.  This
chunking only changes accumulation order.  SGC, MixHop, and H2GCN additionally
use target-node minibatches, with batch size $8$ for Squirrel, $16$ for
Amazon-ratings, and $64$ otherwise; every target still includes its complete
incoming term set.  These cases use minibatch AdamW updates to the same empirical
cross-entropy objective.  The remaining cases use one full-batch update per
epoch.

\subsection{LP implementation and numerical checks}

For every validation and test node, we solve Eq.~\eqref{eq:witness-lp} on the
complete exposed term set without a node or neighborhood cap.  Gurobi uses one
thread, dual simplex (\texttt{Method=1}), disabled presolve, and its default
feasibility and optimality tolerances.  This returns a basic optimum from which
support is measured.  We reconstruct the fixed term directly from checkpoint
logits,
\begin{equation}
  \beta_v=\zbase_v-\sum_a\alpha_{v,a}\ell_{v,a},
  \label{eq:exact-beta}
\end{equation}
which retains all non-adjustable contributions and floating-point accumulation
remainders and enforces $z_v(\alpha_v)=\zbase_v$ exactly.

Numerically, a node is reachable when $\gamma_v(y_v)>10^{-7}$ and a coefficient
is active when $\lambda_{v,a}>10^{-7}$.  TC reach is this event over test nodes;
Err. reach conditions it on errors of the frozen model; Support averages the
active coefficients of reachable solutions.  Head- or stream-specific copies
of one source count as different terms.  No solver failure or violation of the
support bound in Proposition~\ref{prop:witness-region} was observed.

\section{Full Capacity-Control Results}
\label{app:controls}

Every control starts from the same trained checkpoint and train/validation/test
masks as the corresponding repair experiment.  The graph and GNN backbone are
always frozen.  The controls alter only the final prediction rule and are fitted
from training-node labels; validation accuracy selects the checkpoint, and test
labels never affect selection and only score the selected checkpoint.

\paragraph{Refitted linear head.}
Let $r_v\in\R^{d_f}$ be the frozen node embedding immediately before the
original affine classifier.  We standardize each coordinate using its
training-node mean and population standard deviation; coordinates with standard
deviation below $10^{-6}$ are divided by one.  A newly initialized affine map
$W_{\mathrm L}\in\R^{C\times d_f}$, $b_{\mathrm L}\in\R^C$ is then trained on
these standardized embeddings.  No GNN representation, graph coefficient, or
message is updated.  This control asks whether the frozen representation only
needs a different linear boundary; it is intentionally a small linear model,
with $C(d_f+1)$ trainable parameters, rather than a capacity-matched translator.

\paragraph{Parameter-matched node translation.}
The second control retains the original logits and predicts
\begin{equation}
  q_v^{\mathrm N}=4P\tanh\!\left(\rho_{\mathrm N}
  ([h_v,\zbase_v])\right),
  \qquad z_v^{\mathrm N}=\zbase_v+q_v^{\mathrm N},
  \label{eq:node-translation-control}
\end{equation}
where $P$ is the centering matrix from Eq.~\eqref{eq:displacement-set}.
$\rho_{\mathrm N}$ has two affine layers with ReLU and dropout $0.2$ between
them; its final layer is initialized to zero.  Thus it has the same centered,
bounded residual output and exact no-op initialization as set translation, but
it does not receive the message features, their weights, the pooled set vector,
or the total set mass.  The only inputs are the frozen recipient representation
$h_v$ and original logits $\zbase_v$.

For node-control width $H_{\mathrm N}$, its parameter count is
\begin{equation}
  N_{\mathrm N}(H_{\mathrm N})
  =H_{\mathrm N}(d+2C+1)+C,
  \qquad
  H_{\mathrm N}=\arg\min_{h\in\mathbb N}
  |N_{\mathrm N}(h)-N_{\mathrm T}|,
  \label{eq:node-control-matching}
\end{equation}
where $N_{\mathrm T}$ is the actual trainable-parameter count of the
set-conditioned translator in that dataset--backbone case.  The resulting
models contain $101$K--$1.93$M parameters, and the largest relative mismatch in
all $64$ cases is $0.083\%$.  Matching is performed independently for each case
because $d$ and $C$ vary.

Both controls use AdamW with learning rate $10^{-3}$, weight decay $10^{-4}$,
gradient clipping at $5.0$, at most $1100$ epochs, and patience $200$ on strict
validation-accuracy improvement.  This budget is at least as generous as the
set translator's $500$ epochs and patience $80$.  Node translation retains the
frozen prediction as epoch zero, so it cannot be selected below Frozen on the
validation set; the refitted head is selected among its trained affine
checkpoints.  Node translation follows the same target-node batching schedule
as set translation, while the linear head is fitted full batch.

Table~\ref{tab:translation-controls-full} gives the complete matrices
corresponding to the macro comparison in Table~\ref{tab:translation-controls}.
Each cell averages the same ten splits as the main repair table.

\begin{table}[H]
\centering
\caption{Full capacity-control results. Entries are mean test accuracy (\%) over ten splits; rows are backbones and columns are datasets.}
\label{tab:translation-controls-full}
\scriptsize
\setlength{\tabcolsep}{2.6pt}
\renewcommand{\arraystretch}{0.90}
\begin{tabular}{@{}l*{9}{r}@{}}
\toprule
\rowcolor{black!7}\multicolumn{10}{@{}l}{\textbf{(a) Refit head}} \\
Model & Actor & Amazon & Cham. & Cite. & Cora & PubMed & Roman & Squir. & Avg. \\
\midrule
GCN & 28.3 & 44.9 & 67.7 & 76.2 & 87.1 & 88.6 & 51.8 & 51.3 & 62.0 \\
GAT & 28.7 & 44.7 & 64.3 & 75.4 & 87.1 & 87.3 & 45.7 & 49.5 & 60.3 \\
SAGE & 34.8 & 46.2 & 58.9 & 75.6 & 87.5 & 89.2 & 78.4 & 41.9 & 64.1 \\
GATv2 & 28.6 & 46.3 & 65.8 & 75.2 & 87.3 & 87.6 & 63.0 & 52.0 & 63.2 \\
GraphConv & 29.7 & 42.4 & 57.9 & 74.1 & 86.5 & 88.9 & 78.7 & 31.5 & 61.2 \\
SGC & 28.4 & 42.0 & 66.7 & 75.9 & 87.7 & 87.7 & 45.4 & 49.7 & 60.5 \\
MixHop & 32.9 & 41.2 & 56.7 & 76.5 & 86.4 & 89.2 & 76.2 & 39.2 & 62.3 \\
H2GCN & 33.1 & 41.7 & 60.8 & 75.4 & 86.1 & 88.9 & 75.0 & 42.9 & 63.0 \\
\cmidrule(lr){1-10}
Avg. & 30.6 & 43.7 & 62.3 & 75.5 & 87.0 & 88.4 & 64.3 & 44.7 & 62.1 \\
\midrule
\rowcolor{black!7}\multicolumn{10}{@{}l}{\textbf{(b) Node translation}} \\
Model & Actor & Amazon & Cham. & Cite. & Cora & PubMed & Roman & Squir. & Avg. \\
\midrule
GCN & 30.1 & 45.9 & 68.9 & 76.4 & 88.2 & 89.2 & 68.1 & 56.8 & 65.4 \\
GAT & 29.7 & 47.4 & 67.6 & 75.7 & 88.0 & 87.9 & 68.2 & 53.9 & 64.8 \\
SAGE & 35.4 & 47.7 & 61.4 & 75.5 & 87.9 & 89.9 & 79.2 & 42.3 & 64.9 \\
GATv2 & 29.8 & 47.4 & 67.2 & 75.1 & 87.7 & 88.0 & 77.0 & 53.0 & 65.7 \\
GraphConv & 30.0 & 43.6 & 58.5 & 74.9 & 86.9 & 89.3 & 79.6 & 32.8 & 61.9 \\
SGC & 34.8 & 50.0 & 67.6 & 76.0 & 88.5 & 89.8 & 75.8 & 50.0 & 66.6 \\
MixHop & 34.6 & 49.7 & 55.8 & 76.6 & 86.6 & 89.5 & 78.4 & 35.7 & 63.4 \\
H2GCN & 35.1 & 50.1 & 60.1 & 75.2 & 86.8 & 89.5 & 77.5 & 41.0 & 64.4 \\
\cmidrule(lr){1-10}
Avg. & 32.4 & 47.7 & 63.4 & 75.7 & 87.6 & 89.1 & 75.5 & 45.7 & 64.6 \\
\bottomrule
\end{tabular}
\end{table}

\end{document}